\documentclass[10pt,twocolumn,letterpaper]{article}

\usepackage{marvosym}
\usepackage{tabularx}
\usepackage{multirow}
\usepackage{multicol}
\usepackage{wrapfig}
\usepackage{lipsum}
\usepackage[accsupp]{axessibility}
\newcommand{\ModelName}{HunyuanPortrait\xspace}

\usepackage[pagenumbers]{cvpr} 

\definecolor{cvprblue}{rgb}{0.21,0.49,0.74}
\usepackage[pagebackref,breaklinks,colorlinks,allcolors=cvprblue]{hyperref}

\def\paperID{4060} 
\def\confName{CVPR}
\def\confYear{2025}

\title{\ModelName: Implicit Condition Control for Enhanced Portrait Animation}

\author{
Zunnan Xu$^{1,2}$\footnotemark[1]~, 
Zhentao Yu$^{2}$~, 
Zixiang Zhou$^2$~, 
Jun Zhou$^3$\footnotemark[1]~,
Xiaoyu Jin$^1$\footnotemark[1]~,
Fa-ting Hong$^4$\footnotemark[1]~,
\\
Xiaozhong Ji$^2$~, 
Junwei Zhu$^2$~, 
Chengfei Cai$^2$~, 
Shiyu Tang$^2$~, 
Qin Lin$^2$~, 
Xiu Li$^1$\footnotemark[2]~, 
Qinglin Lu$^2$\footnotemark[2]
\\
$^{1}$Tsinghua University \quad  $^{2}$Hunyuan, Tencent \quad 
$^{3}$Sun Yat-Sen Univeristy \quad $^{4}$HKUST \quad
}

\begin{document}

\renewcommand{\thefootnote}{\fnsymbol{footnote}}
\maketitle

\footnotetext[1]{Work done during the internship at Tencent.} 
\footnotetext[2]{Corresponding author.}

\begin{abstract}
We introduce \ModelName, a diffusion-based condition control method that employs implicit representations for highly controllable and lifelike portrait animation. Given a single portrait image as an appearance reference and video clips as driving templates, \ModelName can animate the character in the reference image by the facial expression and head pose of the driving videos. In our framework, we utilize pre-trained encoders to achieve the decoupling of portrait motion information and identity in videos. To do so, implicit representation is adopted to encode motion information and is employed as control signals in the animation phase. 
By leveraging the power of stable video diffusion as the main building block, we carefully design adapter layers to inject control signals into the denoising unet through attention mechanisms. These bring spatial richness of details and temporal consistency. 
\ModelName also exhibits strong generalization performance, which can effectively disentangle appearance and motion under different image styles. Our framework outperforms existing methods, demonstrating superior temporal consistency and controllability. Our
project is available at \href{https://kkakkkka.github.io/HunyuanPortrait}{HunyuanPortrait}.
\end{abstract}

\section{Introduction}
\label{intro}
\begin{figure}[ht]
\centering
\includegraphics[width=\linewidth]{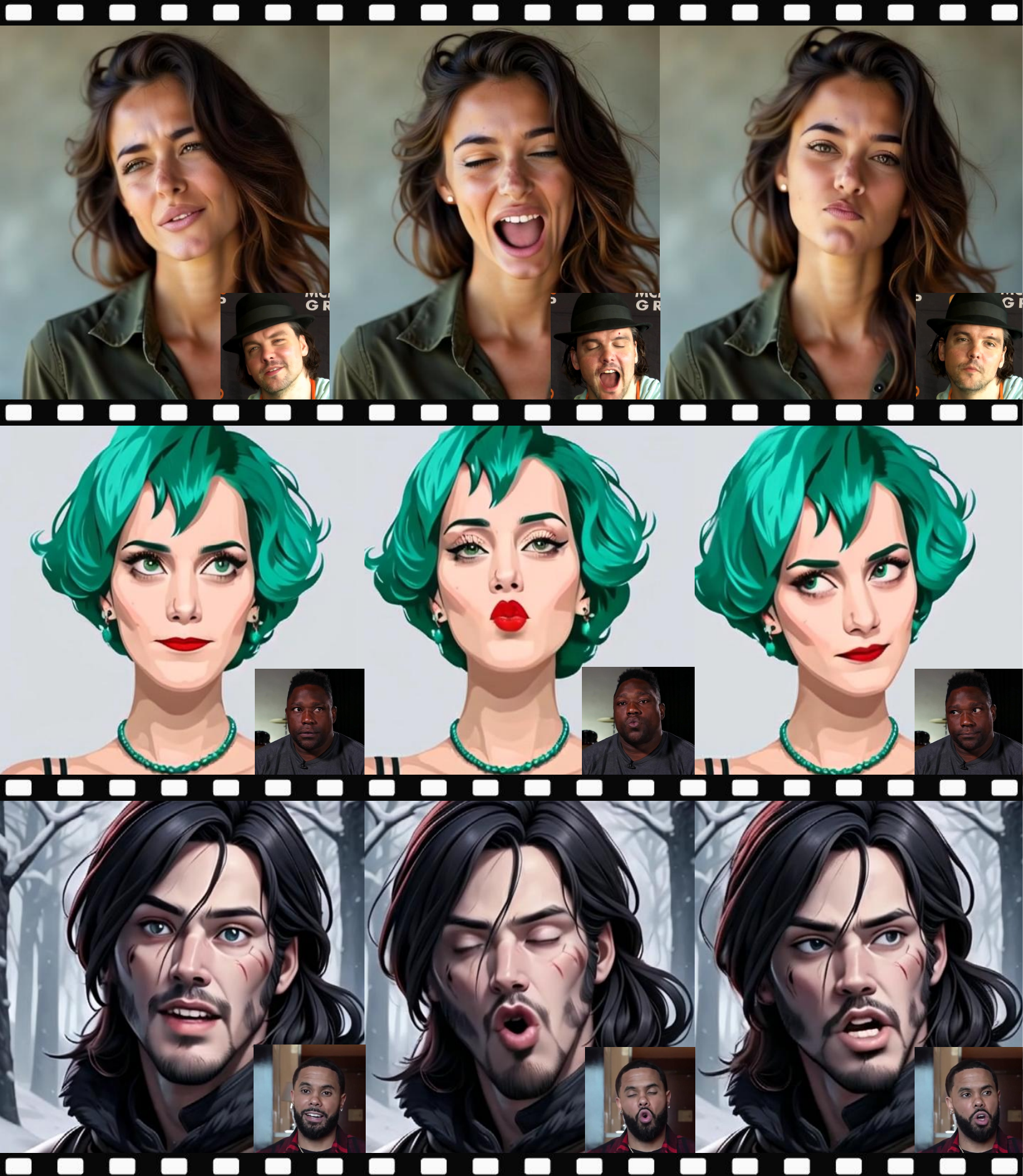}
\caption{Our framework employs implicit condition control to generate portrait animations, demonstrating robust generalization performance with high-fidelity facial dynamics and vivid head poses during cross-reenactment. The animated portraits remain unaffected by variations in facial shape and the spatial position of the driving videos, demonstrating strong identity consistency.}
\label{fig:teaser}
\end{figure}

In this paper, we aim to animate a single static image into a lifelike video using control signals in implicit representations derived from driving videos. As shown in Figure~\ref{fig:teaser}, our primary focus is not only on the high controllability of digital portraits but also on the strong consistency of the characters and background in the reference image.

Portrait animation has been widely applied in fields such as human-computer interaction~\cite{gu2024diffportrait3d,xu2024mambatalk}, virtual reality, and the crafting of digital avatars for the gaming and multimedia industries~\cite{li2023finedance,li2024lodge,li2024lodge++}.
Previous works~\cite{doukas2021headgan,cheng2022videoretalking,zhang2023metaportrait} develop usable portrait animation models utilizing Generative Adversarial Networks (GANs)~\cite{goodfellow2020generative} based on raw video inputs. However, these approaches exhibit limitations in terms of generalization. Given the variability of facial structures and the complexities of facial movements, these methods often face challenges such as artifacts and insufficient control over facial expressions. This is particularly evident when there is a substantial difference in facial shape between the driving video and the source image, resulting in significant issues with facial artifacts and motion distortion.
These early methods often produce results with background jitter and blurred portraits during cross-reenactment~\cite{fomm,nirkin2019fsgan}.

The advancement of diffusion models has led to a significant focus in recent methodologies~\cite{chen2024echomimic,wang2024v} on the fine-tuning of image-level diffusion models~\cite{rombach2021highresolution} through the incorporation of motion modules~\cite{guo2024animatediff}, thereby facilitating the generation of video content.
These methods effectively address the issues associated with GAN-based approaches in managing complex backgrounds and enhancing generalization.
However, even with the incorporation of motion modules, these methods are susceptible to defects during the generation stage (e.g., temporal smoothness and flexibility to different frame rates). This vulnerability arises from the absence of motion pre-training in the underlying models during the pre-training phase.
Additionally, these methods remain constrained to the explicit control of facial expressions.
The effectiveness of these methods, which depend on explicit keypoints to control facial expressions, not only results in the loss of intricate details but also significantly relies on the accuracy of the landmark extraction techniques employed.
Furthermore, as a result of the variability in facial shapes, these methods are limited by post-processing alignment strategies and often face challenges in adapting to diverse facial geometries, which can lead to inaccurate facial control and insufficient preservation of identity in the visual effects of videos.
Due to the human eye's sensitivity to subtle facial changes~\cite{tian2022cues}, it is crucial to preserve the fidelity of facial details and the smoothness of the videos.

Therefore, we propose an implicit conditional control model to tackle the challenges posed by diverse facial geometries and intricate expression details.
We utilize stable video diffusion as the foundational model and integrate identity and motion information through appearance and motion attention. This approach eliminates the need to fine-tune the image diffusion model and to train the motion module separately.
Firstly, we coarsely decouple the identity information from the motion information in the original video by using a pre-trained motion encoder~\cite{drobyshev2023megaportraits}.
Considering that this pre-trained encoder has not completely decoupled identity from motion information, we further implement enhanced training strategies and improved network architectures to strengthen the control of motion features and the separation of portrait identities.
For the adaptation of temporal modeling tasks in video generation using diffusion models, along with the complementary contextual information present between different frames, we propose the incorporation of a motion memory bank to enhance the implicit representation of motion features. 
Due to the substantial variations in the degree of blur and pixel distortion in video frames at different levels of motion intensity, we propose an intensity-aware motion encoder. This encoder aims to improve the capacity of decoupled motion features to capture intricate details.
Additionally, in order to achieve consistent modeling of portrait identity and background areas in videos, we combined ArcFace~\cite{deng2019arcface} with the DiNOv2~\cite{oquab2023dinov2} backbone to develop an enhanced appearance encoder. 
To preserve the pre-training knowledge of both models, we froze their original parameters and proposed an ID-aware Multi-scale adapter (IMAdapter). 
This approach achieves high-fidelity modeling of the intricate details in portraits.
Our main contributions can be summarized as follows:
\begin{itemize}[leftmargin=*,noitemsep,nolistsep]
    \item We propose the first implicit condition animation framework based on stable video diffusion, which significantly improves temporal consistency and achieves high fidelity and precise control for portrait animation.
    \item We develop a series of strategies to enhance identity consistency and improve the accuracy of portrait motion, including data augmentation, enhanced training and inference strategies, and improved network architecture.
    \item Experiments demonstrate the effectiveness of our proposed method, which achieves state-of-the-art performance in both subjective and objective metrics.
\end{itemize}
\begin{figure*}[ht]
\centering
\includegraphics[width=\linewidth]{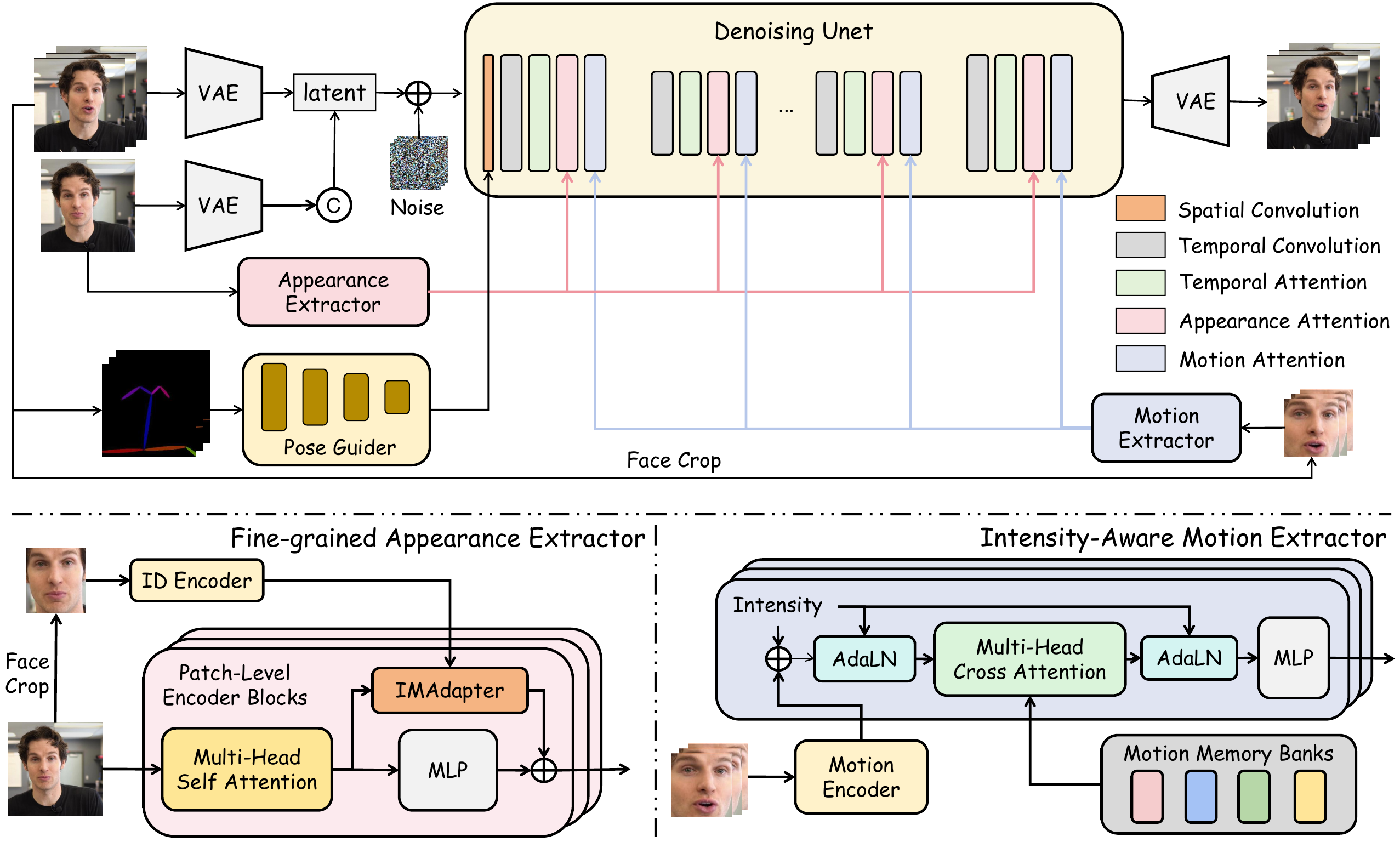}
\caption{Our framework utilizes implicit representation to encode motion information as control signals. By harnessing the capabilities of stable video diffusion as the primary building block, we have meticulously designed a fine-grained appearance extractor to maintain the identity of the portrait, along with an intensity-aware motion extractor to capture intricate facial dynamics. }
\label{fig:pipeline}
\end{figure*}

\section{Related Work}
Human portrait animation aims to bring a still image to life by utilizing a driving source, such as a sequence of facial landmarks or frames that contain the head of the portrait. 
Recent advancements in the field of neural avatars can be divided into two subcategories: non-diffusion-based and diffusion-based methods.
Non-diffusion-based methods~\cite{siarohin2019first,wang2022one,zhao2022thin,hong2023implicit,li2023g2l,zhang2023metaportrait,hong2023dagan++,xu2024vasa,guo2024liveportrait} are renowned for their ability to achieve realistic facial animations and fidelity to motion. 
Many works employed predetermined motion representations that are popular in the literature, such as blendshapes from 3D Morphable Models (3DMM)~\cite{doukas2021headgan,yin2022styleheat,huang2023iddr,yu2023nofa,khakhulin2022realistic,mi2024privacy,tao2024learning}.
However, these works often display significant artifacts when there is a large disparity between the driving portrait and the source image. Due to the use of a warping-based strategy, these methods are also ineffective at capturing videos with significant head and body movements.
With the development of diffusion models~\cite{kong2024hunyuanvideo}, recent methods~\cite{wei2024aniportrait,yang2024megactor,xie2024x,guo2024real} achieve human portrait animation by fine-tuning the stable diffusion model~\cite{rombach2021highresolution}. Several studies have investigated the generation of facial expressions through explicit keypoint control~\cite{wang2024v,zheng2024memo,chen2024echomimic}. 
This method shows great promise for enhancing the quality of generated videos and enables precise control of facial expressions through the extraction of facial landmarks. 
However, given the diversity of facial geometries, there are notable differences in the distribution of facial keypoints among individuals. The alignment processes used in these methods often struggle to manage boundary cases effectively, resulting in identification offsets due to inaccurate alignment.
In addition, the reliance on image-based generative models and separately trained motion modules to achieve temporal consistency results in shortcomings related to temporal smoothness.
In the pursuit of improved temporal continuity, pioneering work~\cite{zhang2024mimicmotion,peng2024controlnext,jin2024alignment,li2024dispose} has employed stable video diffusion models to construct end-to-end diffusion models for human video generation. However, these methods are primarily designed for full-body movements and do not take into account the preservation of facial identity or the diversity of human facial geometries. As a result, they often lead to significant alterations in the identity of the generated face. Additionally, these methods lack the ability to control subtle facial dynamics, which limits their application in generating portraits that incorporate speech content and facial expressions.
In this study, we investigate the strategy of adapting to various face shapes through implicit representations of expression descriptors and achieving capture of facial dynamics. Our method surpasses previous approaches in terms of temporal consistency and generalization ability.

\section{Methodology}
\subsection{Preliminaries}
\noindent\textbf{Diffusion Model.} 
Our framework is built upon diffusion model, as introduced in~\cite{rombach2021highresolution}. The model is designed to facilitate efficient and stable training by conducting the diffusion and denoising processes within a latent space, rather than directly in the image space.
The Variational Autoencoder (VAE), as proposed in~\cite{kingma2013auto}, plays a pivotal role in this process by mapping images from the RGB color space into a latent space. This transformation enables the diffusion process to be directed by textual embeddings.

The process begins with the VAE model, which projects images into latent embeddings. This model comprises two key components: an encoder $\mathcal{E}$ and a decoder $\mathcal{D}$. The encoder $\mathcal{E}$ compresses the pixel space image $x$ into a latent representation $z=\mathcal{E}(x)$, while the decoder $\mathcal{D}$ reconstructs it back into an image space, aiming for $\mathcal{D}(z) \approx x$.
Subsequently, a UNet-based architecture~\cite{ronneberger2015u} is employed for learning the reverse denoising process within the latent space. This network integrates attention mechanisms through Transformer Blocks. The cross-attention mechanism ensures that the textual prompt is effectively integrated throughout the process.
The overall training objective for the UNet can be articulated as follows:
\begin{equation}
\mathcal{L}=\underset{t, z_, \epsilon}{\mathbb{E}} [ \|\epsilon-\epsilon_\theta (\sqrt{\bar{\alpha}_t} z +\sqrt{1-\bar{\alpha}_t} \epsilon, c, t ) \|^2 ],
\label{eq:ldm}
\end{equation}
where $z$ denotes the latent embedding of the training sample. $\epsilon_\theta$ and $\epsilon$ represent the predicted noise by the diffusion model and the ground truth noise at the corresponding timestep $t$, respectively. $c$ is the condition embedding, and the coefficient $\bar{\alpha}_t$ remains consistent with that employed in diffusion models.

\subsection{Overall framework} 
As illustrated in Figure~\ref{fig:pipeline}, building upon stable video diffusion~\cite{blattmann2023stable}, our method is designed with two core components for generating portrait videos: the appearance extractor and the motion extractor.
For the detailed appearance modeling of the portrait, we integrate the appearance encoder to manage the subject's identity and background in the generated video. We carefully design an ID-aware multi-scale adapter based on the Arcface encoder~\cite{deng2019arcface} and DiNOv2~\cite{oquab2023dinov2} backbone to enhance visual appearance representation.
For the motion extractor, we emphasize the enhancement of the implicit representation of facial dynamics. 
By utilizing the pre-trained motion encoder~\cite{megaPortriat}, we further optimize it by taking into account the intensity and the temporal continuity of motions. 
This involved the introduction of motion memory banks and an intensity-aware motion encoder to enhance the perception of motion intensity and refine features.
These representations are conditionally integrated into the denoising U-Net through attention mechanisms.
Our framework also integrates a spatial conditioning signal to ensure the stability of regions beyond the face.
By utilizing these features, we generate latent features using a denoising UNet and subsequently produce video through VAEs.

\subsection{Intensity-Aware Motion Extractor}
For portrait motion modeling, we concentrate on identity-agnostic portrait animation by utilizing an implicit representation of expression descriptors derived from raw videos, which capture a diverse array of facial dynamics (e.g., lip-sync, micro-expressions, eye gaze, and blinking).
This approach differs from existing diffusion-based methods that typically employ explicit landmarks for facial expression representation~\cite{wei2024aniportrait, ma2024follow}.
Considering that the fidelity of facial motion capture is often compromised by extraneous factors—such as variations in facial shape and background noise that can distort essential motion signals—we crop the central area of the face to focus on the regions most indicative of motion dynamics. The boundary is defined as the area between the eyebrows and the bottom of the mouth.
This approach allows us to isolate facial movements and enhance the signal-to-noise ratio.
The cropped pixels from driving videos are subsequently utilized as inputs for the pre-trained motion encoder~\cite{megaPortriat} to extract a coarse implicit conditioning signal.
Since motion intensity varies and exerts distinct influences on the generated pixels, we introduce an Intensity-Aware Motion Encoder  into our framework. This addition enhances the coarse motion features from the motion encoder by adapting to the varying intensities of motion.
The intensity is estimated by two dimensions: the degree of distortion in facial expressions and the overall amplitude of head movement. Considering that the expression and head pose intensity rely on landmark positioning and their relative differences, we propose a calculation method for these variances using landmarks and normalization, which can be formulated as below:
\begin{equation}
\begin{aligned}
e_{k,j} &= l_{k,j} - l_{k,c}, k\in [1,n], j\in [1,m] \\
s &= \sqrt{(l_{1,\max} - l_{1,\min})^2},
\end{aligned}
\end{equation}
where $l_{k,j}$ denotes the j-th landmark point in the k-th frame and $l_{k,c}$ denotes the center of the landmark point (e.g., nose tip). $n$ and $m$ denote the numbers of frames and points. $s$ represents the face scale derived from the first frame of the landmarks. We further calculate the intensity of expressions $I_e$ and the head poses $I_h$ using the formulation below:
\begin{equation}
\begin{aligned}
& I_e = \frac{1}{n\cdot s}\sum_{k=1}^{n}\sqrt{\frac{1}{m}\sum_{j=1}^{m} (e_{k,j} - \frac{1}{n}\sum_{k=1}^{n}{e}_{k,j})^2}, \\
& I_h = \frac{1}{s}\sqrt{\frac{1}{n}\sum_{k=1}^{n}(l_{k,c} - \frac{1}{n}\sum_{k=1}^{n} l_{k,c}})^2.
\end{aligned}
\end{equation}

Considering the issue of accuracy in landmark detection, we discretize the calculated continuous values by dividing the intensity into $d$ discrete levels based on the range of values. We set $d$ to 64 and then map these discrete values using embeddings $\mathcal{D}$ to their corresponding intensity feature vectors $E_{s} = \text{Concat}[\mathcal{D}(I_e),\mathcal{D}(I_h)]$, and concatenate them along the channel dimension to adjust the feature space in response to the dynamic range of motion intensity. In light of the fact that the sequence of motion features is derived from pixel extraction and exhibits a lack of continuity across different contexts, we propose the motion memory bank to improve the context-awareness and temporal modeling capabilities of motion features, which can be formulated as below:
\begin{equation}
\begin{aligned}
&\hat{f}_m = \text{AdaLN}(F_m + E_{s}, E_{s}), \\
&\bar{f}_m = \text{MHCA}(\hat{m}, \hat{f}_m, \hat{f}_m),\\
&\bar{f}_m = \text{AdaLN}(\bar{f}_m, E_{s}), \\
&\bar{f}_m = \text{MLP}(\bar{f}_m), 
\end{aligned}
\end{equation}
where \( F_m \) denotes the features extracted by the pre-trained motion encoder, and \( \hat{m} \) signifies the query features from the motion memory bank, with \( \hat{f}_m \) being utilized as the key and value features. The $\text{MHCA}$ denotes multi-head cross-attention~\cite{vaswani2017attention} and the $\text{AdaLN}$ denotes adaptive layer normalization~\cite{xu2019understanding}.
The learnable memories are designed to supplement contextual information, enriching motion features rather than generating new motions.
This helps the model capture pixel correlation between different frames, alleviating blurring when some facial parts undergo deformation. 
The refined motion features are incorporated into the denoising U-Net using cross-attention mechanisms~\cite{ye2023ip}. 

\subsection{Fine-grained Appearance Extractor}
To achieve appearance control, which encompasses both portrait identity information and background details, we propose an appearance extractor that creates a fine-grained representation of the animated portrait and background.
Existing video diffusion models often struggle with identity consistency~\cite{zhang2024mimicmotion,peng2024controlnext}, which results in deficiencies in the method's capacity to generate portraits.
Due to the human eye's sensitivity to subtle facial changes, visual artifacts can lead to inaccurate facial details and insufficient preservation of identity in the generated videos.
Therefore, we adopt a patch-level image encoder~\cite{oquab2023dinov2} as the backbone of the appearance encoder and introduce an ID-aware multi-scale adapter (IMAdapter) to enhance the model's ID consistency capabilities.
To process the entire image, we first extract information related to clothing and background elements of the portrait. For the enhancement of consistency, as illustrated in Figure~\ref{fig:pipeline}, we utilize the ID encoder to extract identity-related features. We further propose extracting multi-scale, fine-grained identity features using adapters.

\begin{figure}[t]
\centering
\includegraphics[width=0.95\linewidth]{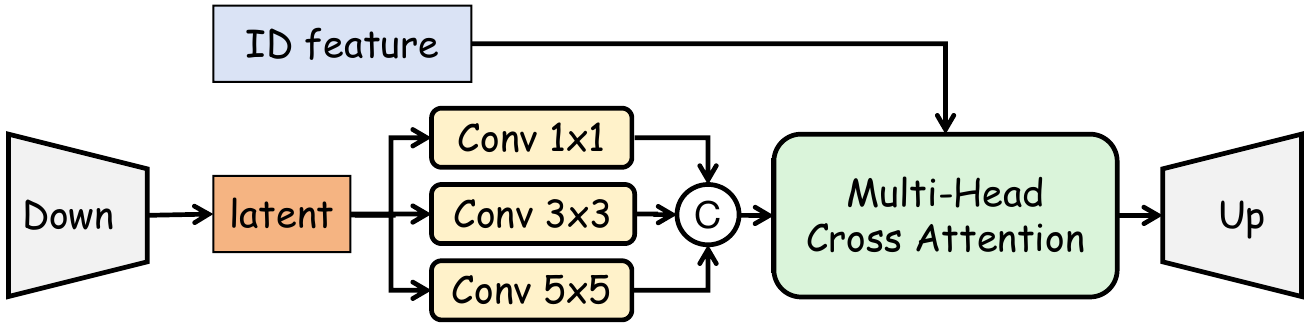}
\caption{The illustration of our ID-aware Multi-scale Adapter (IMAdapter). Here, \textcircled{c} represents the operation of concatenating features along the channel dimension.}
\label{fig:adapter}
\end{figure}

As illustrated in Figure~\ref{fig:adapter}, we first use the linear projection to downsample the visual features. The low-rank features are subsequently fed into convolutional modules at multiple scales, along with a cross-attention mechanism, which captures ID-aware fine-grained details for the preservation of visual identity. The cross-modal features are then linearly projected back to the original dimensions to be merged with patch-level features:
\begin{equation}
\begin{aligned}
&\hat{f}_a = f_a\mathbf{W}_{down}^v, \\
&\hat{f}_c = \text{MConv}(\hat{f}_a), \\
&\bar{f}_a = \text{MHCA}(f_{id}, \hat{f}_c, \hat{f}_c), \\
&f_a = f_a + \bar{f}_a\mathbf{W}_{up}^v,
\end{aligned}
\end{equation}
where \( f_{id} \) represents the ID features, which are utilized as the query features, while \( \hat{f}_c \) serves as the key and value features. \( \text{MConv} \) signifies the multiple convolutions that process appearance features \( f_a \) in parallel and concatenate them along the channel dimension.
During the training phase, the sampling of ID information comes from a random frame in the video sequence; in the inference stage, it comes from the reference image. 

\subsection{Training and Inference Strategies}
For adaptation to various anime styles, we utilize AnimeGANv3~\cite{Liu2024dtgan} to perform style conversion on the dataset. This allows the model to adjust to different styles of driving videos and source images. Instead of directly training models with AnimeGANv3, we employ its style transfer capability to convert facial portrait datasets into anime-stylized representations. These transformed data samples are then integrated into our training dataset for enhanced feature learning.
To mitigate the impact of skin color and other color-related information on facial movements, we employ color jittering to augment the data for the facial cropping input of the motion encoder. 
For maintaining the stability of the regions beyond the face in the generated video, we utilize DWPose~\cite{yang2023effective} as an additional spatial conditioning signal. 
Specifically, we employ the pose-guided network~\cite{hu2024animate} to achieve spatial control condition mapping.
We further incorporate spatial conditional control into the latent features of the first spatial convolutional layer in U-Net through addition.
To simulate scenarios in which certain skeletons from the driving video are absent compared to those in the source image during the inference stage, we implement an augmentation strategy~\cite{wang2024v} for the pose sequence during training. This strategy randomly removes certain edges throughout the training process to improve the stability of spatial control. This mitigates the effects of detector accuracy limitations on detector-based control conditions and offers flexible control capability. During inference, considering that there are differences in the body shape and spatial position of bones between the source image and the driving video, we first translate the position based on the offset of the center point of the face (e.g., the position of the nose tip) and scale it according to the length ratio of the skeletons. We find that the key points of the eyes can lead to subtle variations in face shape. Therefore, in the setting of cross reenactment, we do not utilize the key points and corresponding edges of the eyes.
\section{Experiments}

\begin{table*}[t!]
\centering
\caption{ \textbf{Quantitative results on $512 \times 512$ test images.} LMD multiplied by \( 10^{-3} \), AED multiplied by \( 10^{-2} \), APD multiplied by \( 10^{-3} \) and Identity Similarity multiplied by \( 10^{-1} \). ${\uparrow}$ indicates higher is better. ${\downarrow}$ indicates lower is better. The best results are in bold.}
\resizebox{1.0\linewidth}{!}
{
\begin{tabular}{lcccccc|cccccc}
\toprule
\multirow{3}{*}{Method}  & \multicolumn{6}{c}{\textbf{Self Reenactment}} & \multicolumn{6}{c}{\textbf{Cross Reenactment}} 
\\ 
\cmidrule(lr){2-7}  \cmidrule(lr){8-13}
 & \multirow{2}{*}{\textbf{LMD}\ $\downarrow$} & \multirow{2}{*}{\textbf{FID-VID}\ $\downarrow$} & \multirow{2}{*}{\textbf{FVD}\ $\downarrow$} & \multirow{2}{*}{\textbf{PSNR}\ $\uparrow$} & \multirow{2}{*}{\textbf{SSIM}\ $\uparrow$} & \multirow{2}{*}{\textbf{LPIPS}\ $\downarrow$} & \multirow{2}{*}{\textbf{AED}\ $\downarrow$} & \multirow{2}{*}{\textbf{APD}\ $\downarrow$} & \textbf{Identity}\ $\uparrow$  & \textbf{Facial}\ $\uparrow$ & \textbf{Video}\ $\uparrow$ & \textbf{Temporal} \ $\uparrow$
 \\ 
 & & & & & &  &   &  & \textbf{Similarity} & \textbf{Movements} & \textbf{Quality} & \textbf{Smoothness} \\
\midrule
LivePortrait~\cite{guo2024liveportrait} & 9.14 & 82.71 & 483.38 & 31.41 & 0.72 & 0.22 & 26.83 & 26.97 & 8.71 & 3.00 & 4.11 & 4.06 \\
AniPortrait~\cite{wei2024aniportrait} & 6.67 & 81.90 & 430.24 & 30.54 & 0.67 & 0.27 & 22.42 & 21.32 & 7.95 & 3.83 & 4.00 & 3.81 \\
FollowYE~\cite{ma2024followe} & 5.63 & 77.15 & 417.53 & 30.84 & 0.68 & 0.25 & 20.06 & 20.95 & 8.18 & 4.11 & 4.23 & 3.33 \\
X-Potrait~\cite{xie2024x} & 6.23 & 82.93 & 416.41 & 30.81 & 0.71 & 0.19 & 20.66 & 20.39 & 8.03 & 3.88 & 3.73 & 3.49 \\
\ModelName (Ours) & \textbf{2.02} & \textbf{75.81} & \textbf{333.48} & \textbf{32.98} & \textbf{0.81} & \textbf{0.11} & \textbf{19.45} & \textbf{19.20} & \textbf{8.87} & \textbf{4.55} & \textbf{4.69} & \textbf{4.61}  \\ 
\bottomrule
\end{tabular}
}
\label{tab:metrics}
\end{table*}
\subsection{Experiments Setting}
We train our model using HDTF~\cite{zhang2021flow}, VFHQ~\cite{xie2022vfhq}, VoxCeleb2~\cite{chung2018voxceleb2}, CelebV-HQ~\cite{zhu2022celebv}, and our own collected dataset. This collected dataset comprises monocular camera recordings showcasing 20 hours of authentic human video footage obtained from 200 subjects. The recordings were captured in both indoor and outdoor environments. During the training process, all of the images and videos are resized to $512 \times 512$.
For the optimization of our framework, we employ the AdamW optimizer with a learning rate of \( 1 \times 10^{-5} \). 
The weights of the ID Encoder~\cite{deng2019arcface}, Variational Autoencoder~\cite{kingma2013vae} from Stable Video Diffusion~\cite{blattmann2023svd} and the blocks of DiNOv2~\cite{oquab2023dinov2} are fixed. In the motion memory bank, we incorporate 64 learnable motion memories, each with a dimension of 768, which share parameters across the various blocks. They are initialized using a scaled normal distribution, where each element is sampled from \(\mathcal{N}(0, \frac{1}{\sqrt{\text{hidden\_dim}}})\). 
To ensure stability, we apply gradient norm clipping at a value of 0.99. During inference, we employ the DDIM sampler~\cite{song2020denoising} and set the scale of classifier-free guidance~\cite{ho2022classifier} to 2.0 for our experiment. 
The framework is trained on 128 NVIDIA A100 GPUs for three days.

\subsection{Metrics}
We employ both qualitative and quantitative analyses to assess the generated video quality and motion accuracy of our portrait animation results. For self-reenactment, we utilize the Fréchet Inception Distance (FID)~\cite{heusel2017gans}, the Fréchet Video Distance (FVD)~\cite{unterthiner2019fvd}, 
Peak Signal-to-Noise Ratio (PSNR), Structural Similarity Index (SSIM)~\cite{wang2004image}, Learned Perceptual Image Patch Similarity (LPIPS)~\cite{zhang2018unreasonable} and the Landmark Mean Distances (LMD)~\cite{mediapipe}.
For cross-reenactment, we utilize Average Expression Distance (AED)~\cite{siarohin2019first}, Average Pose Distance (APD)~\cite{siarohin2019first} and the ArcFace Score~\cite{deng2019arcface} as the identity (ID) similarity.
Detailed descriptions of these metrics are provided in \textit{supplementary materials}.
Considering the absence of ground truth and the significant error between current evaluation methods and human perception, we combine the results of user studies that evaluates cross-reenactment from three additional dimensions: facial movements, video quality, and temporal smoothness. The facial movements encompass not only facial expressions but also the overall orientation of the generated portrait head. The quality of the video and its temporal smoothness are primarily assessed by the clarity of the video and the smoothness of the transitions between frames.
For evaluation, we utilize source images and driving videos from two prominent datasets~\cite{chung2018voxceleb2,xie2022vfhq}. This experiment aims to evaluate the model's ability to transfer expressions and movements from one subject to another while preserving the identity of the source portrait.

\begin{figure*}[ht]
  \centering
  \includegraphics[width=\linewidth]{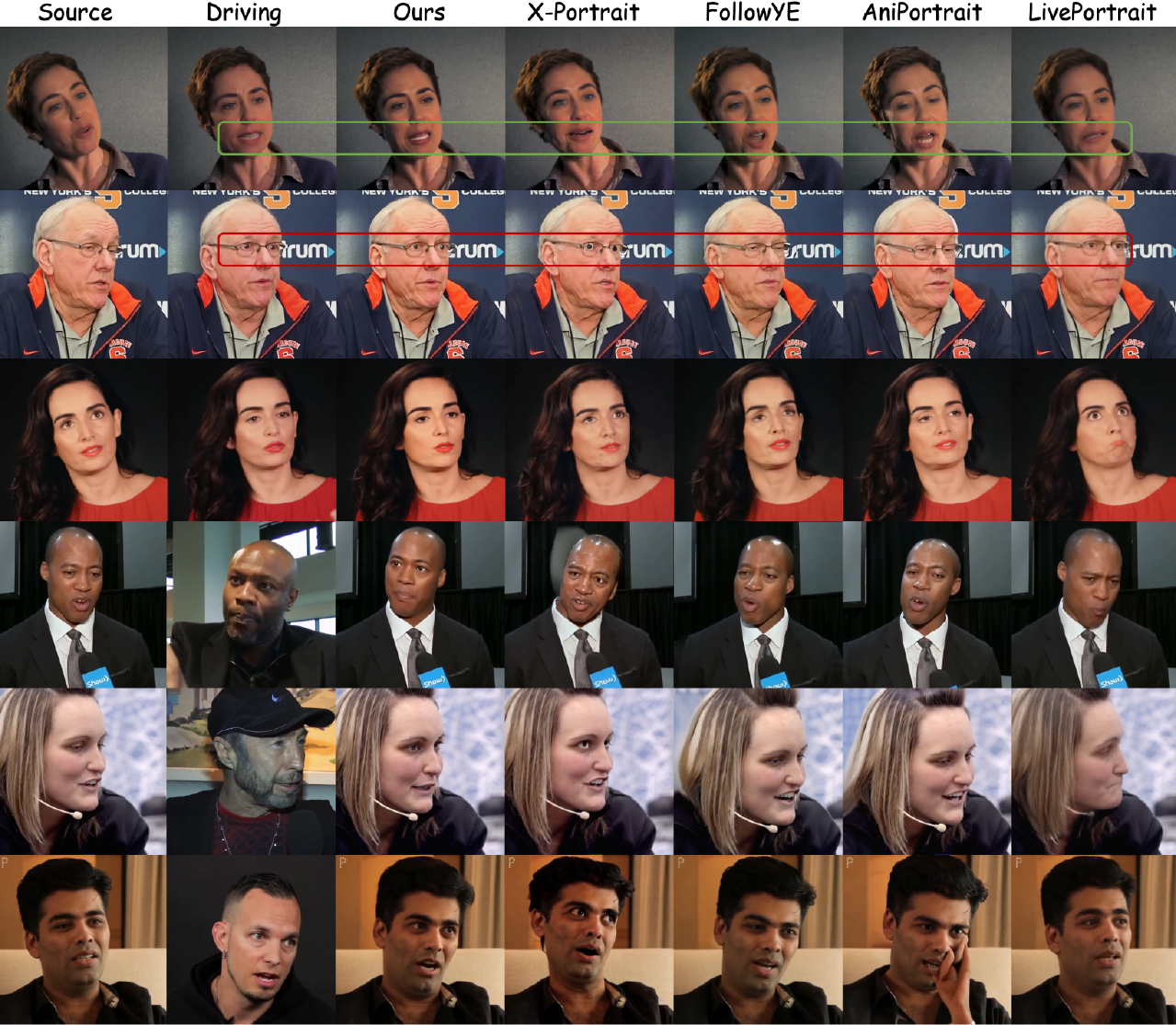} 
  \caption{Qualitative comparisons of self-reenactment and cross-reenactment with state-of-the-art methods.}
    \label{fig:merge}
\end{figure*}

\subsection{Quantitative and Qualitative Analysis}
As shown in Table~\ref{tab:metrics}, our method attains the lowest FID-VID (average FID of frames in videos) and FVD, with a significant advantage when compared to the previously top-performing method, which underscores the superior capability of \ModelName. 
We also find that non-diffusion-based methods have much lower generation quality compared to diffusion-based methods. Although the video resolution is the same, the generated portrait details have significant differences. 
In more challenging cross-reenactment scenarios, our approach demonstrates strong generalization capabilities, leading to substantial improvements across all metrics.
This fully demonstrates the advantage of our method, as cross-reenactment is much more difficult than self-reenactment and is closer to real-life application scenarios.
The warping-based method LivePortrait struggles to accurately model overall head movement, resulting in generated portraits that exhibit minimal variation in head pose relative to the reference image.
Compared with other methods, our method has stronger ID similarity, facial dynamic generation ability, video quality, and temporal smoothness.
Our method employs implicit motion representation rather than explicit facial keypoints to control the generation of facial expressions. 
Our method further takes into account the intensity of motion, the ability to perceive identity, and complementary information based on temporal context. This approach mitigates identity distortion caused by variations in facial geometries and effectively captures dynamic facial details that explicit keypoints cannot represent.
Meanwhile, since our method is based on stable video diffusion, the generated results are smoother compared to previous image-based diffusion methods and have higher video quality. 
These enhancements highlight the superior accuracy and fidelity of our method in capturing fine-grained details, affirming its efficacy in synthesizing realistic portrait motions.
Furthermore, due to the lack of open-source training code, combined with the challenges related to their motion module and the frame rate of the videos, FollowYourEmoji and X-Portrait shows suboptimal performance in terms of temporal smoothness. 
Our method remains unaffected by variations in frame rate.

\subsection{Visualization}
As illustrated in Figure~\ref{fig:merge}, our method can capture more detailed and consistent portrait movements with the driving video.
For self-reenactment, our method generates results that enable more detailed capture of facial dynamics, including eye gaze direction, eye rotation, and lip synchronization.
In addition, our method can also achieve tracking of the overall rotation of the head, which cannot be achieved by previous top-performing warping-based method LivePortrait, such as LivePortrait.
Current state-of-the-art diffusion-based methods have significant defects in maintaining the portrait's identity and facial shape due to the deviation of predicted keypoints during head rotation.
In contrast, our method effectively preserves the identity of the portrait and captures intricate facial dynamics.
Previous methods are influenced by the differences in facial geometries between the driving video and the reference image, leading to substantial challenges in preserving the identity of the generated video.
This is a result of inadequate disentanglement of their appearance features and facial dynamics.
These issues result in problems such as blurry videos and distorted portraits.
Our framework effectively addresses the disentanglement of the two key elements involved in portrait animation, thereby demonstrating a strong generalization ability, as shown in Figure~\ref{fig:merge}.
The diffusion-based method exhibits superior generation quality; however, discrepancies in facial keypoints due to variations in facial shape, combined with the limited ability of these keypoints to accurately capture dynamic facial details, make the previous state-of-the-art method less effective than ours in generating features such as facial geometry, eye gaze direction, and lip synchronization in portraits.
Due to variations in facial shapes, methods that rely on explicit keypoints are influenced by the facial characteristics of individuals in the driving video. This leads to insufficient separation between motion and identity.
Furthermore, since their approach utilizes the image generation model Stable Diffusion as its foundation, it also exhibits inferior performance compared to ours in terms of temporal smoothness.
Building upon stable video diffusion and utilizing our meticulously designed fine-grained feature extractors, our method exhibits exceptional performance in identity preservation, video quality enhancement, and the capture of subtle facial movements, resulting in more vivid and smoother generated videos.

\subsection{Ablation Study}

\begin{figure}[t]
\centering
\includegraphics[width=\linewidth]{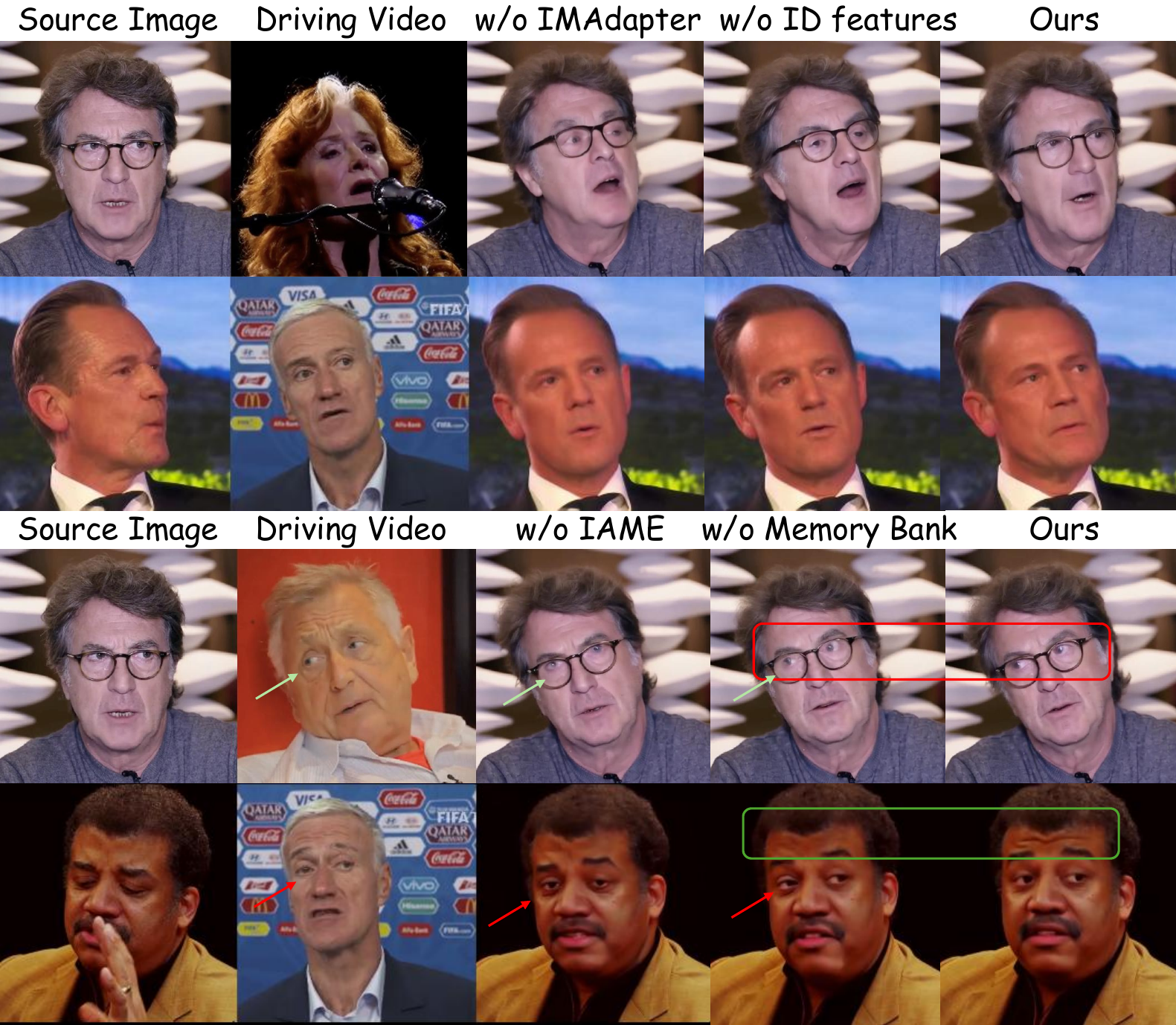}
\caption{Qualitative ablation studies of different components.}
\label{fig:ablat}
\end{figure}

\noindent\textbf{Effect of Motion Memory Bank.}
We validate the effectiveness of the motion memory bank. As shown in Table~\ref{tab:abla}, the incorporation of the motion memory bank results in higher-quality videos, demonstrating improvements in FID-VID and FVD, as well as enhanced facial movements. We also observe an improvement in smoothness. Through the utilization of motion memories, our methodology demonstrates context-based perception capabilities, thereby enhancing motion features in the presence of ambiguous information. By integrating memories with information from preceding and subsequent frames, the memory bank can provide more detailed information into facial dynamics. As illustrated in Figure~\ref{fig:ablat}, the forehead wrinkles associated with the elevation of the eyes become more pronounced, when the character's eyebrows are raised.

\begin{table}[t]
\centering
\caption{Ablation study to validate the effectiveness of different components. LMD (Landmark Mean Distance) multiplied by $ 10^{-3} $ and Identity Similarity multiplied by $ 10^{-1} $.}
\resizebox{1.0\linewidth}{!}
{
\begin{tabular}{lcccc}
\toprule
\multirow{2}{*}{Setting}  
 & \textbf{Identity}\ $\uparrow$  & \multirow{2}{*}{\textbf{FID-VID}\ $\downarrow$} & \multirow{2}{*}{\textbf{FVD}\ $\downarrow$} & \multirow{2}{*}{\textbf{LMD}\ $\downarrow$}\\
 & \textbf{Similarity} & & & \\
\midrule
\ModelName       & 8.87 & 75.81 & 333.48 & 2.02 \\ 
- Memory Bank    & 8.75 & 78.43 & 361.94 & 2.78 \\ 
- IAME           & 8.63 & 80.79 & 385.43 & 4.01 \\ 
- ID Features    & 8.21 & 75.03 & 330.12 & 1.98 \\ 
- IMAdapter      & 8.09 & 77.14 & 352.67 & 2.54 \\ 
\bottomrule
\end{tabular}
}
\label{tab:abla}
\end{table}

\noindent\textbf{Effect of Intensity-Aware Motion Encoder (IAME).}
We confirm the significant role of the intensity-aware motion encoder. As illustrated in Table~\ref{tab:abla}, the incorporation of this component, in contrast to the direct utilization of coarse motion features derived from the pre-trained motion encoder~\cite{megaPortriat}, offers critical guidance and intricate details regarding the facial dynamics associated with portrait motions. This enhancement significantly elevates the overall quality of the generated video, evidenced by an improvement in FID-VID, FVD, and ID similarity. As illustrated in Figure~\ref{fig:ablat}, the absence of feature refinement in the IAME may result in inaccuracies in the micro-expressions of the characters, including the direction of their gaze. Furthermore, the intricate details of the face are neglected.

\noindent\textbf{Effect of ID features and IMAdapter.}
To validate the effectiveness of the IMAdapter, we first remove the ID features from the IMAdapter, leaving only multiple convolution kernels for low-rank feature processing.
As shown in Table~\ref{tab:abla} and Figure~\ref{fig:ablat}, the ablation results demonstrate that incorporating ID features is critial in enhancing the model's ID consistency ability.
We further remove IMAdapter from the appearance extractor's blocks. As demonstrated in Figure~\ref{fig:ablat}, the incorporation of IMAdapter provides more fine-grained features related to facial texture, geometry, and other facial characteristics.

\section{Conclusion}
In this paper, we present an enhanced portrait animation framework that addresses the limitations of previous approaches. Our model leverages implicit representations with stable video diffusion to achieve temporal consistency and precise control over facial dynamics.
The proposed model demonstrates significant improvements in the overall quality and temporal consistency of generated videos, allowing for more natural and lifelike animations.
We offer a robust framework for generating high-quality, controllable animations, which achieve high disentanglement. Our method not only enhances the visual experience but also opens up possibilities for immersive content creation.
With continued research and development, this approach holds great potential for applications in virtual reality, gaming, and human-computer interaction.
However, as with any technology that generates realistic human likenesses, future work should develop safeguards to ensure responsible use.

{
    \small
    \bibliographystyle{ieeenat_fullname}
    \bibliography{main}
}

\newpage
\appendix

\section{Benchmark Metrics Details}
We employ both qualitative and quantitative analyses to assess the generated video quality and motion accuracy of our portrait animation results. 
In evaluating self-reenactment, we consider multiple metrics: the Peak Signal-to-Noise Ratio (PSNR), the Structural Similarity Index (SSIM) ~\cite{wang2004image}, and the Learned Perceptual Image Patch Similarity (LPIPS) ~\cite{zhang2018unreasonable}. Specifically, for the LPIPS metric, we apply the AlexNet-based perceptual similarity measure LPIPS~\cite{zhang2018unreasonable} to gauge the perceptual similarity between the generated animated images and the driving images.
We also utilize the Fréchet Inception Distance (FID)~\cite{heusel2017gans} to assess image quality, the Fréchet Video Distance (FVD)~\cite{unterthiner2019fvd} to evaluate temporal consistency, and the Landmark Mean Distances (LMD) to measure the accuracy of generated facial expressions.
The landmarks are extracted using Mediapipe~\cite{mediapipe}. We compute the average Euclidean distance between the facial landmarks~\cite{mediapipe} of the reference and generated frames. Lower values indicate better geometric accuracy.
FID~\cite{heusel2017gans} is utilized to measure the similarity in feature distribution between generated and real images, employing Inception-v3 features. Lower scores indicate better perceptual quality. Additionally, the FVD is used to evaluate temporal coherence through features extracted from a pretrained network~\cite{unterthiner2019fvd}.
For cross-reenactment, we utilize the ArcFace Score~\cite{deng2019arcface} as the identity (ID) similarity metric between the generated frames and the reference image.
For Average Expression Distance (AED)~\cite{siarohin2019first} and Average Pose Distance (APD)~\cite{siarohin2019first}, we calculates the Manhattan distance of expression and pose parameters from SMIRK~\cite{retsinas20243d}, with lower values indicating better expression and pose simiarity.

\section{Discussions, Limitations and Future work}
In this section, we make comparisons of our method on ID modules with X-Portrait \cite{xie2024x} and DiffPortrait3D \cite{gu2024diffportrait3d}. The ID modules share the common objective of preserving the identity of the reference portrait during the generation of new animations or views. However, each method employs distinct strategies to achieve this goal, leading to differences in implementation, training, and application focus.

Our method utilizes a fine-grained appearance extractor coupled with an ID-aware multi-scale adapter (IMAdapter). This design enables detailed modeling of identity and background information from the reference image, ensuring high-fidelity identity preservation in the generated animations. The IMAdapter incorporates multi-scale convolutions and cross-attention mechanisms, which enhance the model's ability to capture intricate identity features and maintain their consistency across different frames and contexts.
In contrast, X-Portrait's ID module extracts appearance and background features from a single reference image, which are then concatenated into the UNet's transformer blocks. This straightforward approach ensures consistent identity representation across generated frames but may not capture fine-grained details as effectively as our multi-scale architecture. X-Portrait focuses on expressive portrait animation, aiming to transfer facial expressions and head poses from driving videos to the reference portrait while maintaining identity similarity.

DiffPortrait3D adopts a different strategy by injecting appearance context from the reference image into the self-attention layers of a frozen UNet. This method effectively preserves identity across various rendering views but is primarily designed for 3D view synthesis rather than full animation. It leverages the generative power of pre-trained diffusion models to synthesize 3D-consistent novel views from as few as a single portrait.

For future work, we believe that exploring the synthesis of images from unknown perspectives to enhance identity retention capability is crucial. The current method has limitations in maintaining identity consistency from unknown perspectives after significant head rotation. The ID preservation design of DiffPortrait3D presents a potential improvement; however, it is still constrained by specific angles and cannot achieve the generation of unknown 360-degree views. We assert that enhancing the identity retention capability from unknown perspectives is a vital and feasible direction for improving the method proposed in this paper.

Besides, there are still several limitations with our method. Currently, our methodology is restricted to generating only the head and shoulder portions of portraits. 
We try to apply our method for generating full-body portraits that include hands; however, the results are unsatisfactory. The generated images of hands occasionally exhibit deformities and blurriness.
This limitation stems from the inadequate representation of hand regions within the dataset, which hinders our ability to accurately render hand details. 
Future work can enhance our method by incorporating data that includes hand movements and improving the representation of hand features, thereby extending our approach to encompass full-body movements. 
Additionally, our approach is limited by the inherent constraints of the diffusion model. 
The significant computational costs impede the real-time applicability of our methods. Future work can accelerate the generation process through model distillation.

\section{More Implementation Details}
\subsection{Appearance Extractor}
For the input of the appearance extractor, we first resize the reference image to 256x256.
We use the DiNOv2-Large with 4 register tokens as the appearance extractor, and fix the weights of its backbone during training. 
In the implementation of the IMAdapter, we first reduce the dimensionality of the features to 384 using linear layers. Subsequently, we employ convolutional kernels of varying scales (e.g., $1\times 1, 3\times 3, 5\times 5$) for parallel processing and fuse these with ID features through a multi-head cross-attention mechanism. We set the number of heads for the multi-head cross-attention to 8.

\subsection{Motion Extractor}
The motion extractor comprises a total of six blocks within the network, utilizing consistent network configurations. The dimensionality of the latent features is set at 768, with the attention mechanism employing eight heads. The activation function implemented is the Sigmoid Linear Unit (SiLU)~\cite{elfwing2018sigmoid}. 
The motion encoder is borrowed from MegaPortrait~\cite{megaPortriat}.
We upgrade the architecture of the motion encoder from ResNet-18 to ResNet-50 to facilitate the pre-training of the motion encoder.
After completing pre-training, the weights of this motion encoder is fixed and utilized for fine-tuning SVD.

\section{More Visualizations}
\begin{figure*}[htbp]
\centering
\includegraphics[width=.96\linewidth]{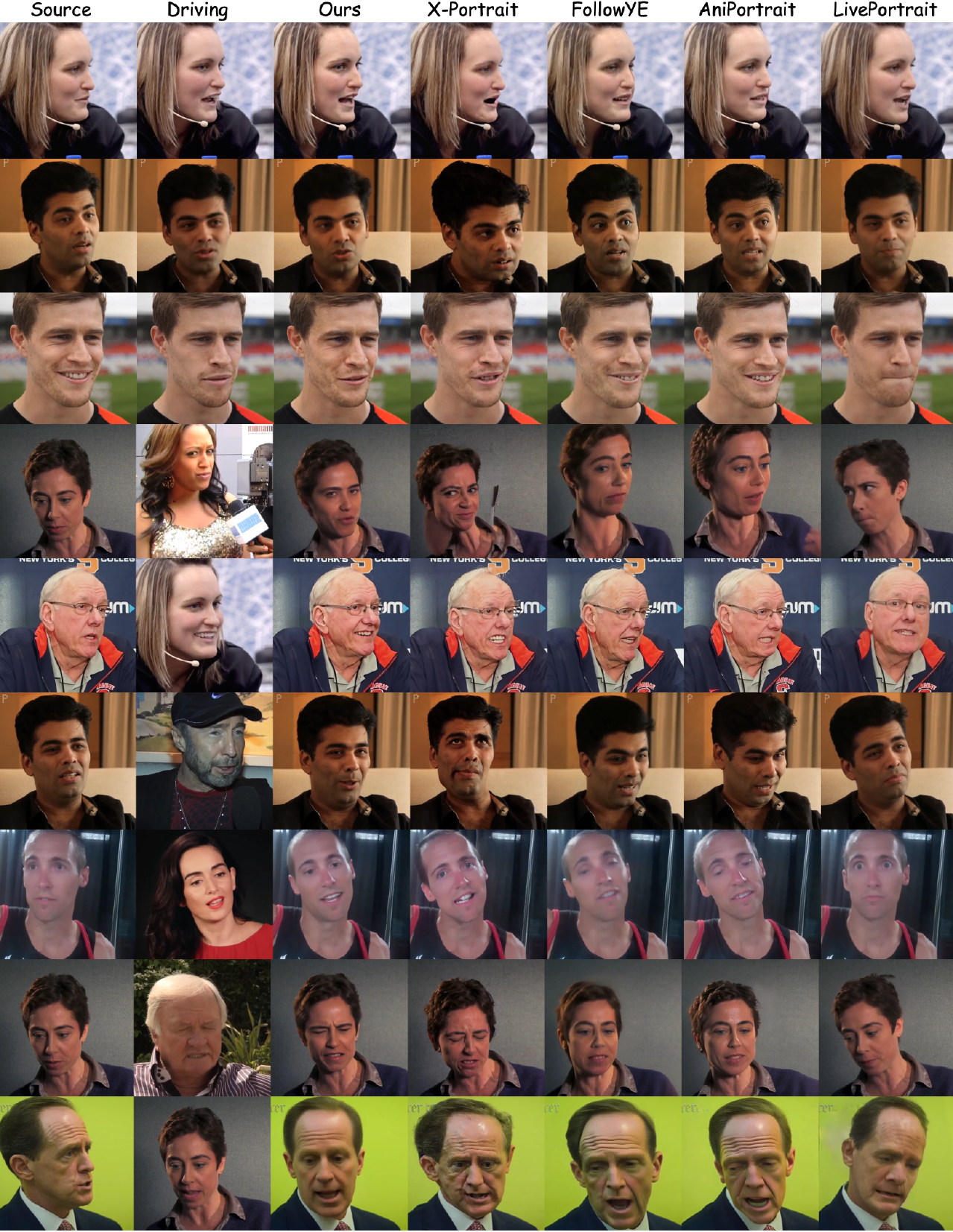}
\caption{More visualizations of self-reenactment and cross-reenactment.}
\label{fig:merge2}
\end{figure*}

\begin{figure*}[htbp]
\centering
\includegraphics[width=0.8\linewidth]{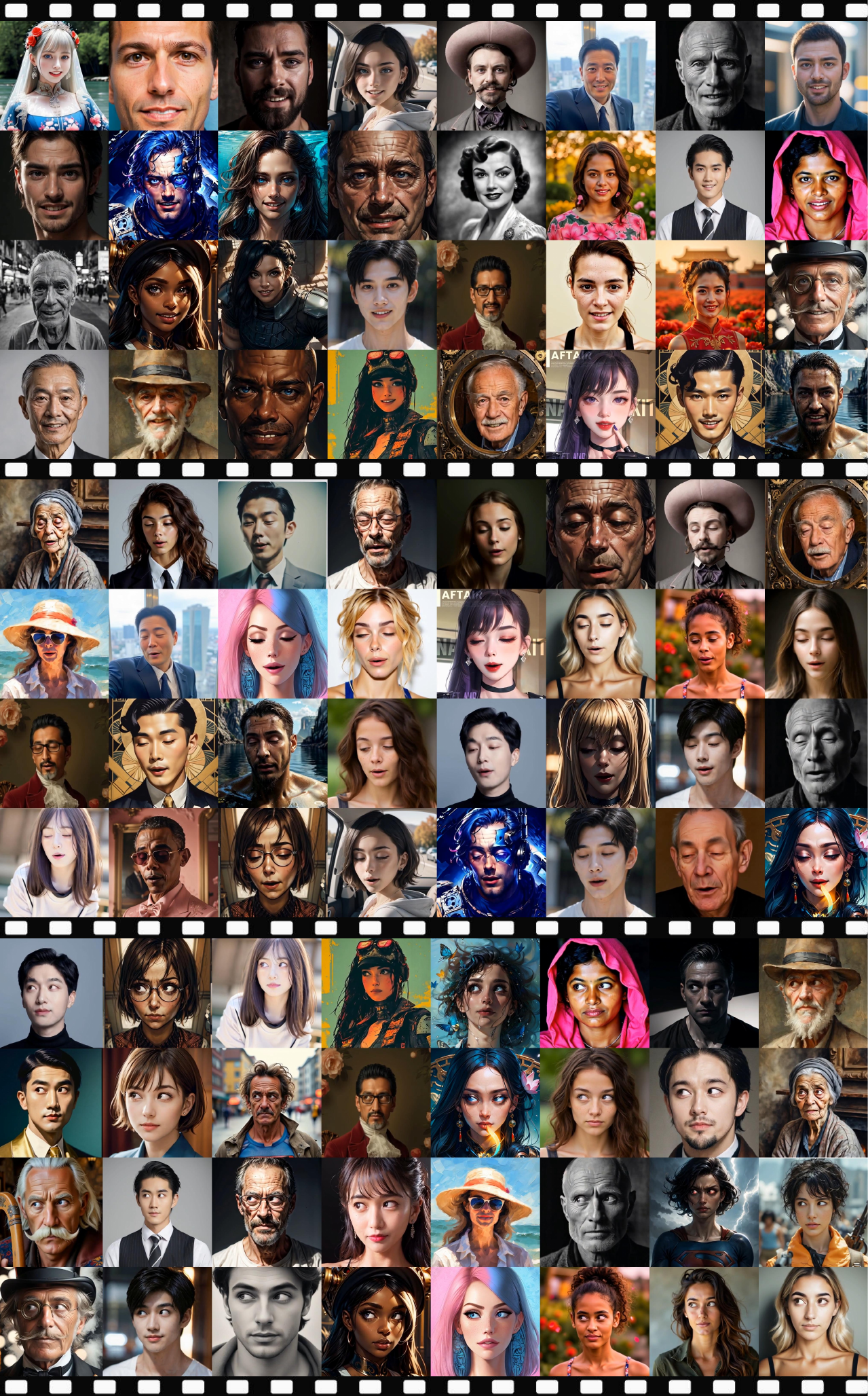}
\caption{More visualizations of animated portraits.}
\label{fig:sup1}
\end{figure*}

\begin{figure*}[htbp]
\centering
\includegraphics[width=0.8\linewidth]{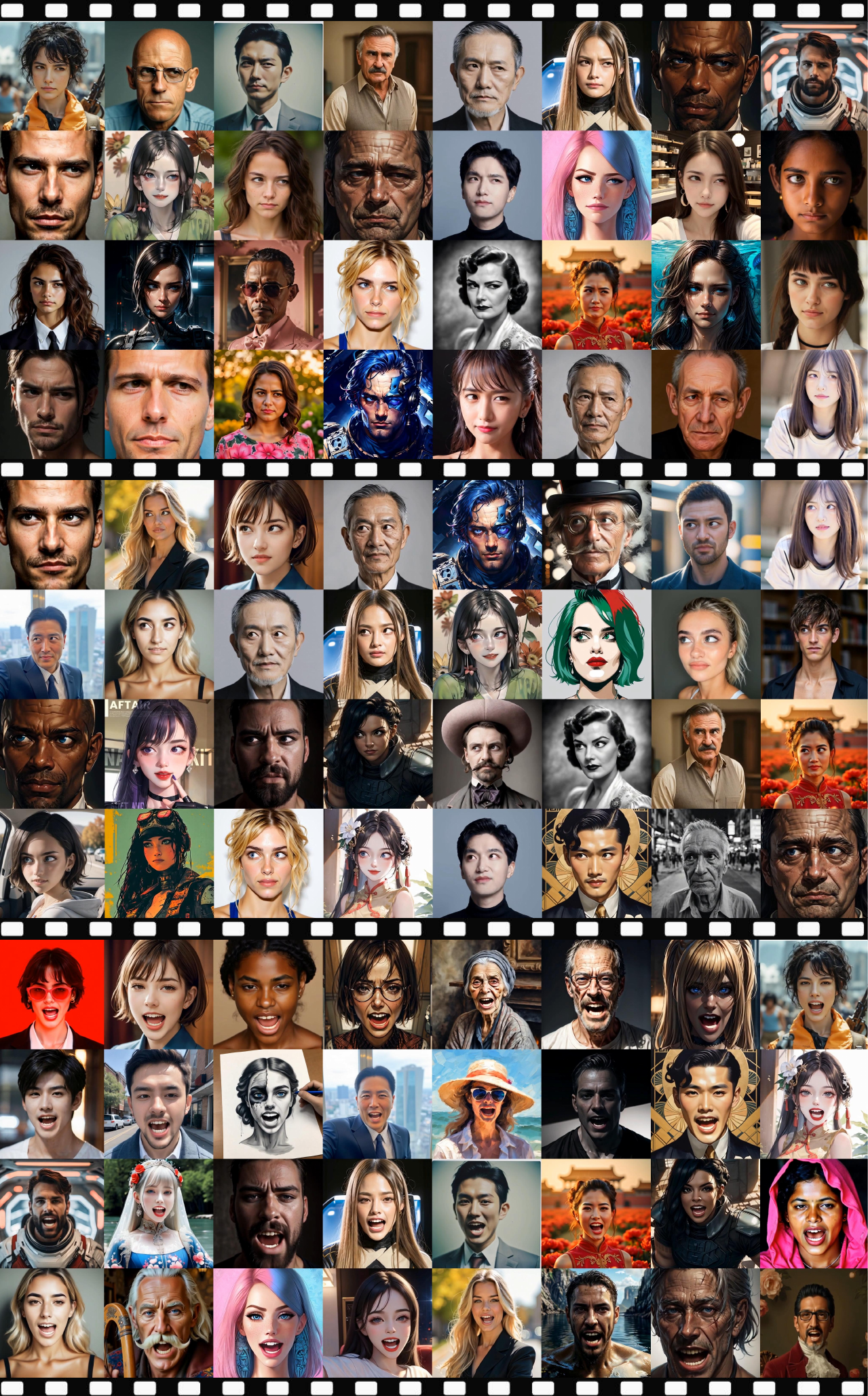}
\caption{More visualizations of animated portraits.}
\label{fig:sup2}
\end{figure*}

\begin{figure*}[htbp]
\centering
\includegraphics[width=0.8\linewidth]{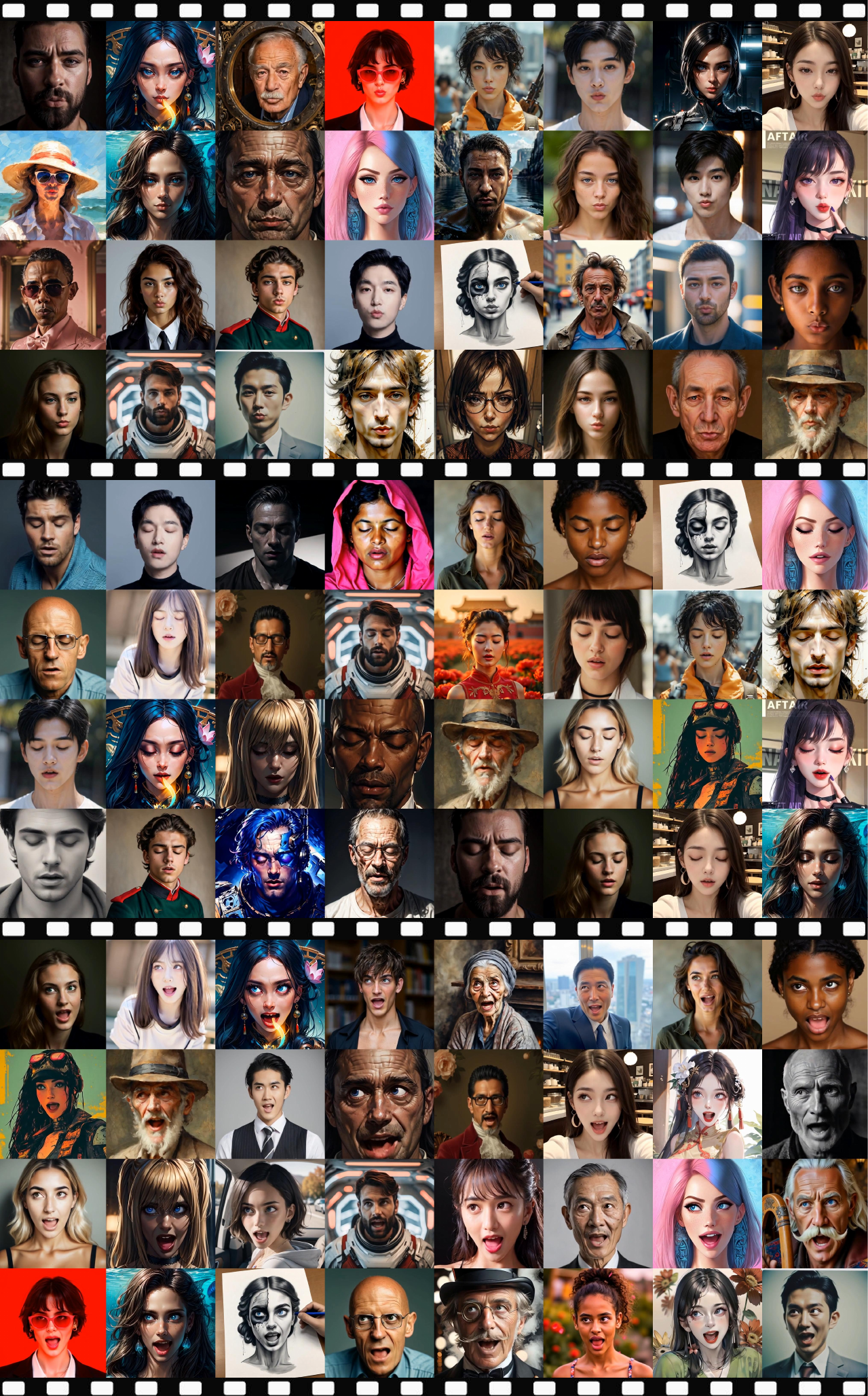}
\caption{More visualizations of animated portraits.}
\label{fig:sup3}
\end{figure*}

As shown in Figure~\ref{fig:merge2}, we present additional visualization results under the self-reenactment and cross-reenactment setting to better demonstrate the decoupling capability of our method in terms of appearance and motion. 
In order to demonstrate the robust generalization of our method, we selected a variety of images and videos from different style domains, including Civitai~\footnote{\url{https://civitai.com}}, Bilibili~\footnote{\url{https://bilibili.com}}, and VFHQ~\footnote{\url{https://liangbinxie.github.io/projects/vfhq}}, for demonstration purposes.
To avoid copyright issues with driving videos, we first use a source image along with the original video to obtain the generated result. 
As illustrated in Figure~\ref{fig:sup1}, Figure~\ref{fig:sup2} and Figure~\ref{fig:sup3}, we then utilize the generated results as the driving videos to animate other videos.

\section{Ethics Consideration}
\subsection{User Study Details}
\begin{figure*}[htbp]
\centering
\includegraphics[width=\linewidth]{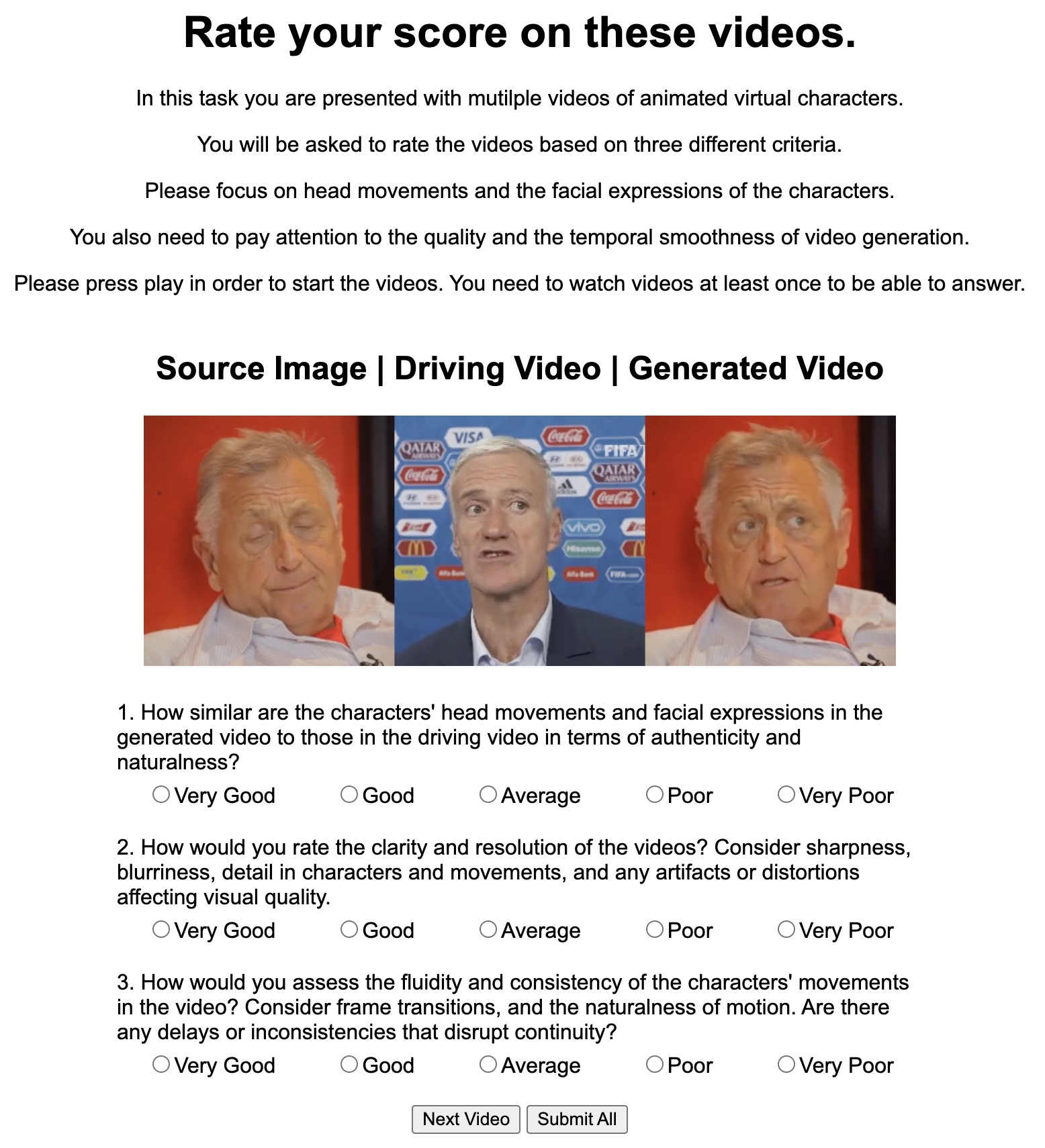}
\caption{The screenshots of user study website for participants.}
\label{fig:study}
\end{figure*}

In the user study, a total of 120 experienced participants are invited to take part. 
For each participant, we paid compensation that exceeded the local average hourly wage.
We employ three metrics: Facial Movement, Video Quality, and Temporal Smoothness. For each metric, participants are presented with a video rated on a five-point scale. The grading options available to participants were as follows: Very Good (5), Good (4), Average (3), Poor (2), and Very Poor (1). As illustrated in Figure~\ref{fig:study}, the online evaluations are conducted using a well-structured website questionnaire. The questionnaire provides a comprehensive guideline, along with several example videos at the beginning. These example videos are not included in the rating but serve to illustrate the quality of video generation and ensure consistency in the rating criteria among participants.

\subsection{Societal Impacts and Responsible AI}
Our focus is on advancing the visual effects of virtual AI avatars to enhance their effectiveness for beneficial applications. 
It is essential to clarify that our research objectives are not intended to deceive or mislead. 
Like other content creation methods, our approach is not immune to potential misuse. 
We firmly oppose any misuse that could result in the creation of deceptive or harmful content through the impersonation of real individuals. 
Despite the risks of misuse, it's important to highlight the substantial positive outcomes of our technology. 
These include promoting educational fairness, aiding those with communication difficulties, and offering companionship or therapeutic aid. 
The significance of our research is underscored by its potential to assist those in need. 
We are committed to the ethical progression of AI, with the goal of fostering human welfare.
The output videos generated by our method still retain identifiable traces of the actual individuals they are based on. 
To mitigate the potential for abuse, we are developing a neural network-powered tool designed to differentiate between genuine and synthetic videos, which includes our synthetic talking face videos in the training dataset. 
We will keep the community updated on any advancements in our models.

\end{document}